\documentclass[sigconf,natbib=false]{acmart}

\AtBeginDocument{%
  \providecommand\BibTeX{{%
    \normalfont B\kern-0.5em{\scshape i\kern-0.25em b}\kern-0.8em\TeX}}}

\begin{document}

\title{Prototype of deployment of Federated Learning with IoT devices}

\author{Pablo García Santaclara}
\affiliation{%
  \institution{I\&C Lab. atlanTTic research centre. University of Vigo}
  \city{Vigo}     \country{Spain}
  \postcode{36310}
}

\author{Ana Fernández Vilas}
\authornotemark[1]
\affiliation{%
  \institution{I\&C Lab. atlanTTic research centre. University of Vigo}
  \city{Vigo}     \country{Spain}
}
\email{avilas@det.uvigo.es}
\orcid{1234-5678-9012}

\author{Rebeca P. Díaz Redondo}
\affiliation{%
  \institution{I\&C Lab. atlanTTic research centre. University of Vigo}
  \city{Vigo}     \country{Spain}
}
\email{rebeca@det.uvigo.es}
\orcid{1234-5678-9012}

\begin{abstract}
In the age of technology, data is an increasingly important resource. This importance
is growing in the field of Artificial Intelligence (AI), where sub fields such as Machine Learning (ML) need more and more data to achieve better results. Internet of Things (IoT) is the connection of sensors and smart objects to collect and exchange data, in addition to achieving many other tasks.   A huge amount of the resource desired, data, is stored in mobile devices, sensors and other Internet of Things (IoT) devices, but remains there due to data protection restrictions. At the same time these devices do not have enough data or computational capacity to train good models. Moreover, transmitting, storing and processing all this data on a centralised server is problematic. Federated Learning (FL) provides an innovative solution that allows devices to learn in a collaborative way. More importantly, it accomplishes this without violating data protection laws. FL is currently growing, and there are several solutions that implement it. This article presents a prototype of a FL solution where the IoT devices used were raspberry pi boards. The results compare the performance of a solution of this type with those obtained in traditional approaches. In addition, the FL solution performance was tested in a hostile environment. A convolutional neural network (CNN) and a image data set were used. The results show the feasibility and usability of these techniques, although in many cases they do not reach the performance of traditional approaches.
\end{abstract}


\maketitle
\section{Introduction}

The \textbf{Internet of Things (IoT)} paradigm is composed of sensors and smart objects interconnected to collect and exchange data, take actions, automate different tasks, over the Internet and other communication protocols. The data generated by all these devices can come from many parts of the world and must be put to good use. IoT devices have endless uses and applications, of which a few of the most worth mentioning include: (1) Industry; where collected data provide insight into productivity and efficiency so that different aspects of the production chain can be improved., e.g.  machine utilization, speeding up improvement, etc. (2)  Smart cities, where IoT play a central role in areas such as parking management,  healthcare monitoring, waste management, etc \cite{saleem2020building}; (3) Smart home, where the integration of different devices, appliances and sensors within a house increases comfort and efficiency in  energy consumption; (4) Smart grid, where sensors  can prevent failure points, extend component life and optimise costs.

According to  \textit IoT Analytics \footnote{iot-analytics.com}  the number of connected IoT devices in Spring 2022 exceed 12 billion worldwide, growing by 9\% this year, despite all the supply problems  due to the pandemic and other issues. This number is set to grow dramatically by 2025, surpassing 25 billion connected IoT devices by then. It is difficult to estimate the amount of data generated by all these IoT devices but it is quite clear that the numbers are massive. All this generated data is very valuable, but with the increasing number of IoT devices, it does not scale well to analyse all of them with centralised solutions. At these sizes, storage capacity, processing and even transmission become challenges. Achieving the processing capacity for gigantic amounts of data on a centralised server is very costly. In addition, a centralised model compromises data security and privacy.

There is a growing common view that the transition from centralized ML to distributed ML at the network edge is necessary, but largely complex, for a number of reasons \cite{journals/pieee/ParkSBD19}. Most prominent among these is the disconnection between current principles of network practice (coding, link communications, random access, protocol assumptions) and the way ML algorithms are designed and analyzed, with ideal trustful agents \cite{8970161}. While it is generally agreed that the intelligence of ML should be moved closer to the devices (data producers located at the network edge) and benefit from plentiful computing nodes, the emerging design of efficient distributed ML algorithms has to deal explicitly with the heterogeneity of the computing and communication equipment (e.g, from IoT sensors to cloud servers; from wireless channels and strong interference to local data; from privacy concerns to public data) \cite{9252927}.

A first breakthrough for employing multiple nodes for training and guaranteeing privacy is federated learning, which enables model synthesis out from a large corpus of decentralized data \cite{10221843}. The Federated Learning approach is based on training on devices with their own data, these devices share their models, which are aggregated on a central coordinating server. This way, the devices do not have to transmit their data to a server at any time. This technique also makes it possible to share the computational tasks among many devices. Since having to train a model on a server with extremely large amounts of data is becoming an increasingly frequent problem for big companies. This technology must be flexible in its use, as most possible usage scenarios contemplate problems in communications or in the availability of clients. A case where this is easy to visualise is an implementation with smartphones, where it is very likely that some of them will lack internet connection or even battery power. Therefore, the performance of Federated Learning for cases where communication is unstable should be tested.

The main contribution of this work is introducing an implementation architecture for FL in IoT scenarios. This implementation combines the Amazon Cloud with and edge layer consisting on restricted Raspberry  devices. For this implementation technologies, we report our results under ideal conditions  where the devices in the edge are always working and the communication is considered reliable and, in the otter hand, some hostile configurations. The results allows to extract conclusions about the performance of FL on IoT scenarios. More sophisticated FL approaches such as collaborative Federated Learning are being researched in the literature to achieve a more realistic approach to the privacy, security communication and reliability conditions of machine learning for  IoT.   The paper is organised as follows. Section \ref{sec:background} establishes a context for Federated Learning and Section \ref{sec:scenario} is a brief overview of the technological basis of out IoT and it  outlines the different parts of the architecture for the experiments. Then, Section \ref{sec:experiment} describes the conditions under which these experiments were performed.  In Section \ref{sec:results}, the results obtained after conducting the experiments are discussed. Finally, the paper is ended with a conclusion in Section \ref{sec:conclusions}.

\section{Background}
\label{sec:background}

The concept of \textbf{Federated Learning} was first proposed by Google in 2016 \cite{konecny}. Originally, their intention was to alleviate the problem of having too much data for a single node, where storing the whole dataset on a single node became unfeasible. His proposal was to build the ML models based on datasets that were distributed across multiple devices. This at the same time satisfied the concern that large enterprises have in recent years to improve data security and user privacy, as decentralising data fulfils the function of making data leakage more difficult \cite{yang2019federated}. Conventionally, if a set of data owners set out to train an ML model, they would gather all the data together to then train the model. In a system with Federated Learning, the owners collaboratively train a joint model, each training the model with their own data and then exchanging the results with the other owners, improving together a global model. This aims to achieve a similar accuracy to the one originally described without exposing each owner's data to the others. 

In 2017 McMahan et al. \cite{mcmahan2017communication} presented two main concepts \textbf{FederatedSGD(FedSGD)} and \textbf{FederatedAveraging(FedAVG)}. In FedSGD, a fraction C of the customers in each round is chosen, the initial model is communicated to these and the model is trained by each client, then the average model is calculated. A typical FedSGD implementation is one with C=1 and a fixed learning coefficient. In which each client trains with its own data the current global model, communicates the model obtained to the server and the server is in charge of aggregating them and updating the global model. A solution, as described above, is known as Federated Averaging. The amount of computation done in each round is controlled by 3 parameters; C, the number of clients in each round. E, the number of training iterations each client does in each round. B, the size of the local minibatch used in client updates.

\begin{figure*}[htp]
  \centering   \includegraphics[width=0.8\textwidth]{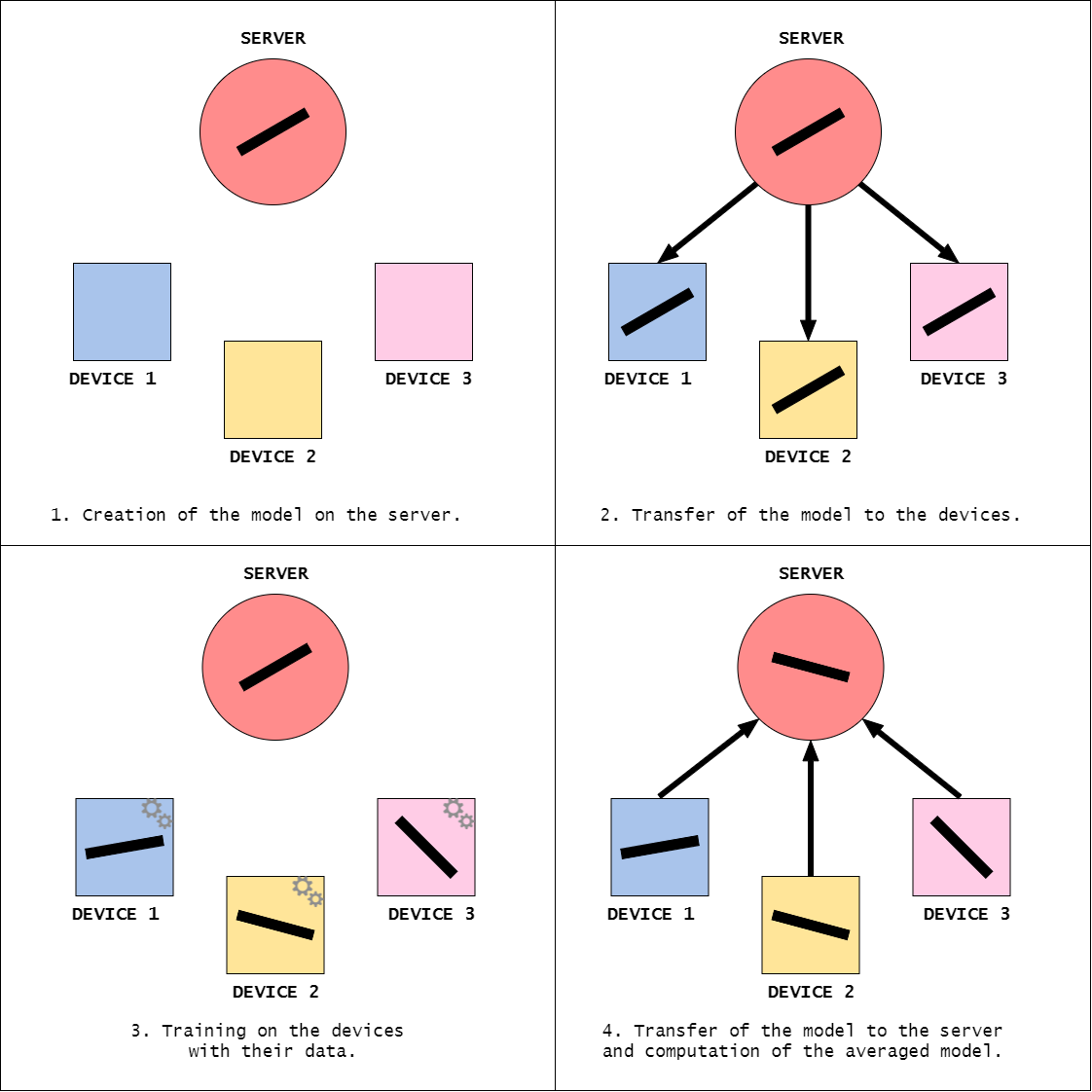}
  \caption{Federated Learning steps}   \label{fig:FL}
\end{figure*}

Therefore, generally, each iteration of a Federated Learning algorithm consists of the following steps. An illustration describing these steps can be found in the \autoref{fig:FL}: (1)selected sample of customers downloads the current global model; (2) Each of these selected clients computes an updated model based on the data they possess; (3) The models updated by the clients are sent to the server; (4) At the server, all these updated models are aggregated to improve the global model.

%% ventajas en entornos con conectividad no garantizada
Due to its characteristics, its use is advantageous in environments where connectivity is not guaranteed, especially if the number of clients is significant. Since, even if a client is disconnected at a given point in time, due to the functioning of Federated Learning, when it becomes available again, it will pick up the global model, drawing on the data of all those involved in the communication. This means that the sudden addition of new clients will not lead to a deterioration of the global model. As the whole process could also be anonymous, new clients could join in to further improve the model, without this being a problem.

The use of systems with Federated Learning still has some challenges to overcome. The main one is heterogeneity in systems and data. Clients may have very different computational, storage or communication capabilities, and devices may experience power shortages or connectivity problems during an iteration. Moreover, it cannot be assumed that the data on each client is IID (independent and identically distributed) \cite{zhao2018federated}, which complicates model training tasks. Communication can also become a problem, being a potential bottleneck. To mitigate the possibility of it becoming a drawback, communication efficiency can be improved, mainly by reducing the size of the messages transmitted in each round.

%\textbf{buscar otros trabajos que estén relacionado con el uso de FL en entornos %IoT y presentar una visión general.  Parte del TEXTO ABAJO ESTÁ SACADO DE %COMPROMISE}

%%%%%%%%%%%%%%%%%%%%%%%%%%%%%%%%%%%%%%%%%%%%%%%%%%%%%%%%%%%%%%%%%%%%

\section{Scenario}
\label{sec:scenario}
\begin{figure*}[h]
  \centering
  \includegraphics[width=0.8\linewidth]{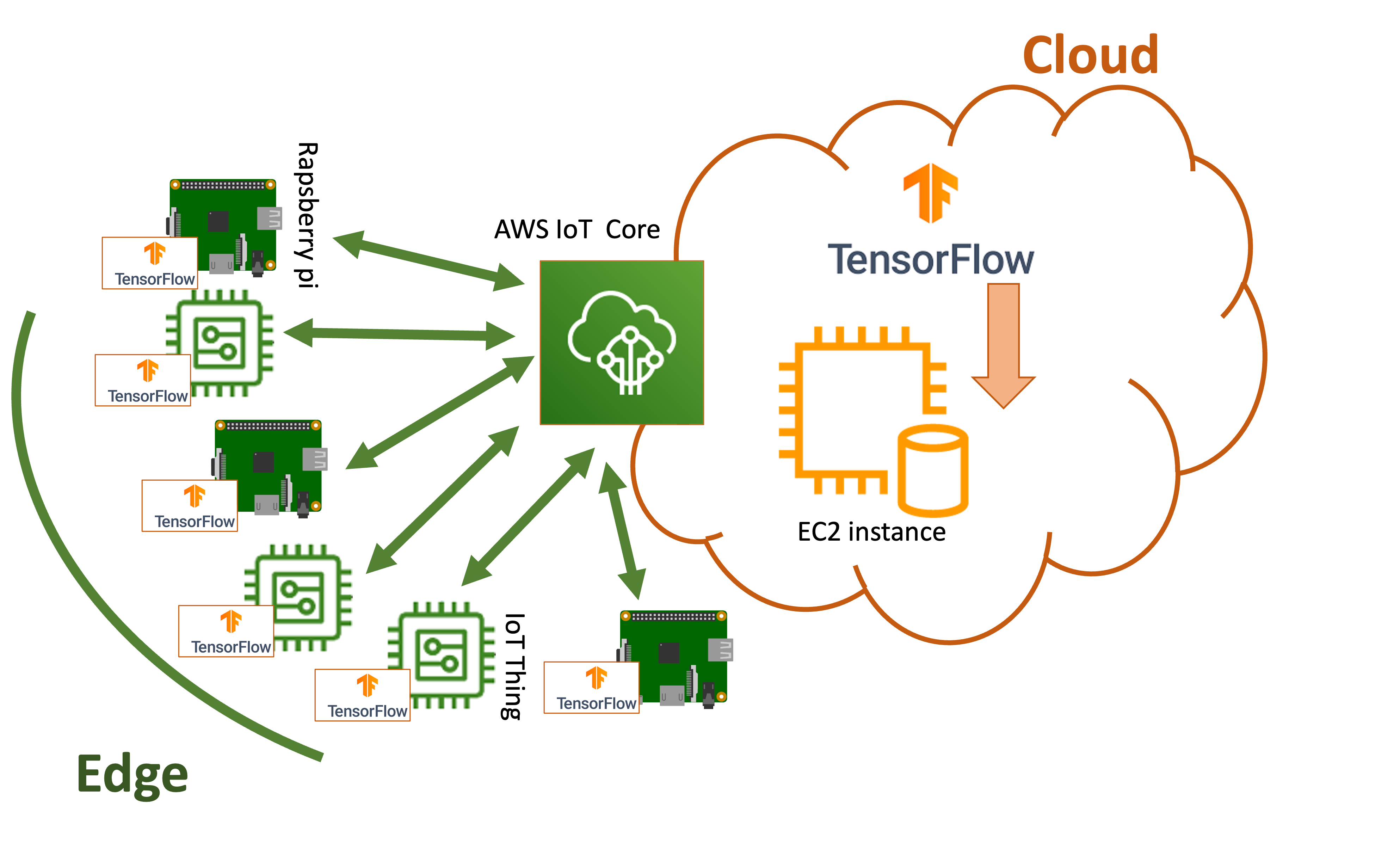}
  \caption{System architecture.}
  \label{fig:arch}
\end{figure*}

 Federated Learning uses the Edge Computing paradigm for its operation. Edge Computing is closely related to IoT technologies. It is based on bringing the processing and data storage closer to where the data is being generated, the clients in this case. Therefore, when intelligence is figured out in a Edge computing environment, Federated Learning turns into the natural solution. We propose and scenario where federated learning and edge computing technologies are integrated into  the architecture  in \autoref{fig:arch}. The Edge is materialised as a set of edge devices, Raspberry in our implementation; this edge layer in integrated into a public cloud, Amazon cloud. At the cloud level, an EC2 \cite{ec2} instance, a virtual server in Amazon's Elastic Compute Cloud, plays the role of FL aggregator or central server. It will be responsible for creating the initial model and distributing it to the edge devices. It will then wait to receive the trained models from the clients, create the global model and redistribute it, so that everyone improves jointly.

 From an architectonic point of view, the integration of the edge layer into the cloud relies on the usage of AWS IoT Core, and AWS feature that enables IoT devices to connect to the AWS Cloud \cite{IOTCORE}. It is responsible for registering devices, acting as the device registrar. It also acts as the gateway to the Cloud, in addition to authorisations and being the messaging broker. Every raspberry used was registered as a "thing" in AWS IoT, in order to be able to communicate with the Core, which manages communications. AWS IoT Core plays the role of message broker between the edge layer and the FL aggregator by using MQTT as communication protocol, and would allow the models to be transmitted back and forth.  
 
From a ML point of view, both edge devices and the aggregator will execute ML processes  in order to create, train and evaluate the models. TensorFlow \cite{abadi2016tensorflow}, an open source library mainly used for Machine Learning, was the library used to perform all of the aforementioned operations. On the server, it was used to create the initial model and to evaluate the performance of the global model. On the clients, it was required to train the model.

%%%%%%%%%%%%%%%%%%%%%%%%%%%%%%%%%%%%%%%%%%%%%%%%%%%%%%%%%%%%%%%%%%%%

\section{Experiment}
\label{sec:experiment}
Any Federated Learning solution requires clients with their own data to train their models, and a central server that coordinates communications and which is responsible for weighting the models of the clients and distributing the global model. For our experiment, it is necessary to set up some edge devices to act as clients. Providing them with data so that they can train their respective models. Consequently, a dataset will also have to be obtained in order to distribute its data among all the clients.

\begin{figure}[h]
  \centering
  \includegraphics[width=\columnwidth]{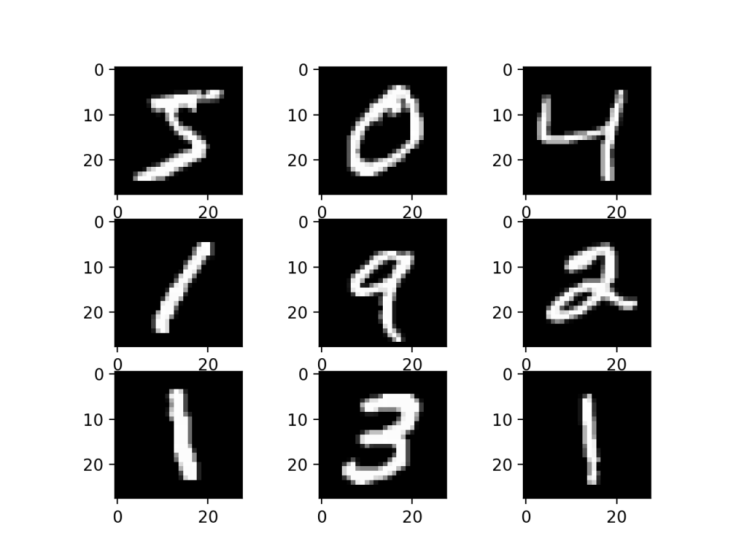}
  \caption{MNIST dataset samples.}
  \label{fig:mnist}
\end{figure}

The choice of the dataset and model to be used had to take into account the processing and dynamic memory capacity of the raspberrys. The model could not be too complex so that the raspberrys would not be able to train it. Besides, the dataset would have to be extensive so that it could be divided among several clients. That choice was to take the MNIST dataset \cite{lecun-mnisthandwrittendigit-2010}. This is a dataset made up of handwritten numbers by different people. The training set consists of 60,000 samples while the test set has 10,000 samples. The images are made up of 28x28 pixels, they look as shown in \autoref{fig:mnist}. This dataset is well known and is common for people who are learning Machine Learning techniques. \par 

%%%%%%%%%%%%%%%%%%%%%%%%%%%%%%%%%%%%%%%%%%

Raspberry boards were chosen to simulate the devices that would act as clients. Five boards were used for the experiment, three of them Raspberry Pi 3 Model B whereas the other two were Raspberry Pi 2 Model B. All of them have 1 Gb of RAM, this plus CPU capacity supposed a limiting factor. Their description is as follows:
\begin{itemize}
    \item  Raspberry Pi 3 Model B \cite{RPI3}: CPU (Broadcom BCM2387 64bit ARMv7 Quad Core 1.2GHz); RAM (1GB LPDDR2; Wifi (Yes)
    \item   Raspberry Pi 2 Model B \cite{RPI2}: CPU (Broadcom BCM2836 900MHz quad-core ARM Cortex-A7); RAM (1GB LPDDR2); Wifi (No)
\end{itemize}

An AWS instance, as mentioned above, simulates the central server, with the following technical characteristics:  Model t2micro;  vCPU 1; Mem 8 GiB.

The model used to be trained with this dataset is a simple convolutional neural network, which are known for performing well in image classification tasks. Keras \cite{chollet2015keras}, a deep learning library, running on top of TensorFlow. is used to create the model. The model looks as shown in \autoref{tab:Model}, it consists of an input layer, two hidden layers and an output layer. The input layer flattens the input data, then a regular dense layer that uses \textit{ReLU} as activation function and a dropout layer which helps prevent overfitting, finally the output layer is another regular dense layer which uses \textit{Softmax} as activation function. \textit{ReLU} activation function is a linear function that will output the input directly if it is positive, or zero otherwise. \textit{Softmax} activation function converts a vector of numbers into a vector of probabilities.

\begin{table}[h] \centering
\begin{tabular}{c|c}
 \hline   \textbf{Layer (type)} & \textbf{Output Shape}  \\
 \hline   Flatten  & 784  \\
  \hline   Dense & 128 \\
 \hline   Dropout  & 128  \\
 \hline   Dense  & 10  \\
\hline
\end{tabular} \caption{CNN model.}\label{tab:Model}
\end{table}

 The list of weights is passed as a bytearray in the message payload via MQTT communication. In the receiver, the model is reconstructed with the keras \cite{chollet2015keras} method \textit{loadmodel()}. On each raspberry pi board, 4 clients were simulated using multiprocessing, where each process is a client. Making a total of 20 clients participating in Federated Learning. This was the maximum number of participants in the different tests.

%%%%%%%%%%%%%%%%%%%%%%%%%%%%%%%%%%%%%%%%%%
\section{Results}
\label{sec:results}
We have deployed different experiments where the focus was put on the distinctive characteristics of IoT environments in terms of reliability of tiny devices and communication issues. So, we consider a friendly environment \ref{subsec:friendly} where IoT devices do not fail and communications from the edge to the central aggregator and back is always possible. After that, different hostile scenarios are considered in section \ref{subsec:hostile},  

\subsection{Friendly environment}
\label{subsec:friendly}

As mentioned in the previous section, we use the MNIST one dataset. It follows a quite uniform distribution of samples, which can be seen in the \autoref{tab:mnist}. In its totality it has 60,000 samples for training and 10,000 for testing. The choice was that each client would have 300 training samples, as a lower number would not give as well-representative results, and a larger number would start to give problems to the Raspberry devices. On the other hand, the 10,000 test samples are always used, which is necessary so that the results of the different tests can be compared.

\begin{table}[h] \centering
\begin{tabular}{|l||*{5}{c|}}\hline
Dataset &
label 0 & label 1 & label 2 &label 3 & label 4\\\hline\hline
Train &5,923&6,742&5,958&6,131&5,842\\\hline
Test &980&1,135&1,032&1,010&982\\\hline
\end{tabular}
\begin{tabular}{|l||*{5}{c|}}\hline
Dataset &
label 5 & label 6& label 7& label 8 & label 9\\\hline\hline
Train &5,421&5,918&6,265&5,851&5,949\\\hline
Test &892&958&1,028&974&1,009\\\hline
\end{tabular}  \caption{Distribution of MNIST samples}\label{tab:mnist}
\end{table}

The objective of the model trained is to predict which number is the one in the image that is introduced. There are different metrics to measure performance.  In order to  asses the performance, we use the regular measures: (1)  \textit{Accuracy}, which is the number of correct predictions divided by the number of total predictions, this metric works best when the number of samples of each label is the same. MNIST is close to having this equality; (2)\textit{Confusion matrix}, a matrix showing the number of False positives, False negatives, True positives and True negatives; (3)  \textit{F1 score}, which is the harmonic mean between precision and recall, and seeks a compromise between this two; (4)  \textit{MAE} and \textit{MSE}, which aim to give an average of the distance between predicted and actual values, this is not useful for this classification problem, since similar numbers are for example 1 and 7, but their distance would not express anything; and (5) \textit{Loss}, which is not a metric, but is used by the neural network when training, being the distance between real and predicted values. Being what the neural network seeks to minimise during training.

The decision made was to use accuracy and loss. Accuracy is one of the most universal metrics and allows to easily know the performance. Loss, on the other hand, relates well to precision, as combined they allow us to know what is happening. For example, if both increase, it could be due to overfitting, i.e. it will adjust to learning the particular cases we teach it and will be unable to recognise new data.  However, if the accuracy increases while the loss decreases, it is assumable that the neural network is learning correctly.

The second important point to consider is how to evaluate the performance of Federated Learning. The results were compared with those obtained in a traditional architecture, where data is centralised. Therefore, the performance of Federated Learning would be compared with the results that a client would obtain with only its own data, in case data restrictions prevented them from being shared, and on the other hand, with the results that a centralised server would obtain with the data of all the clients.

\begin{figure*}[h]
  \includegraphics[width=\textwidth]{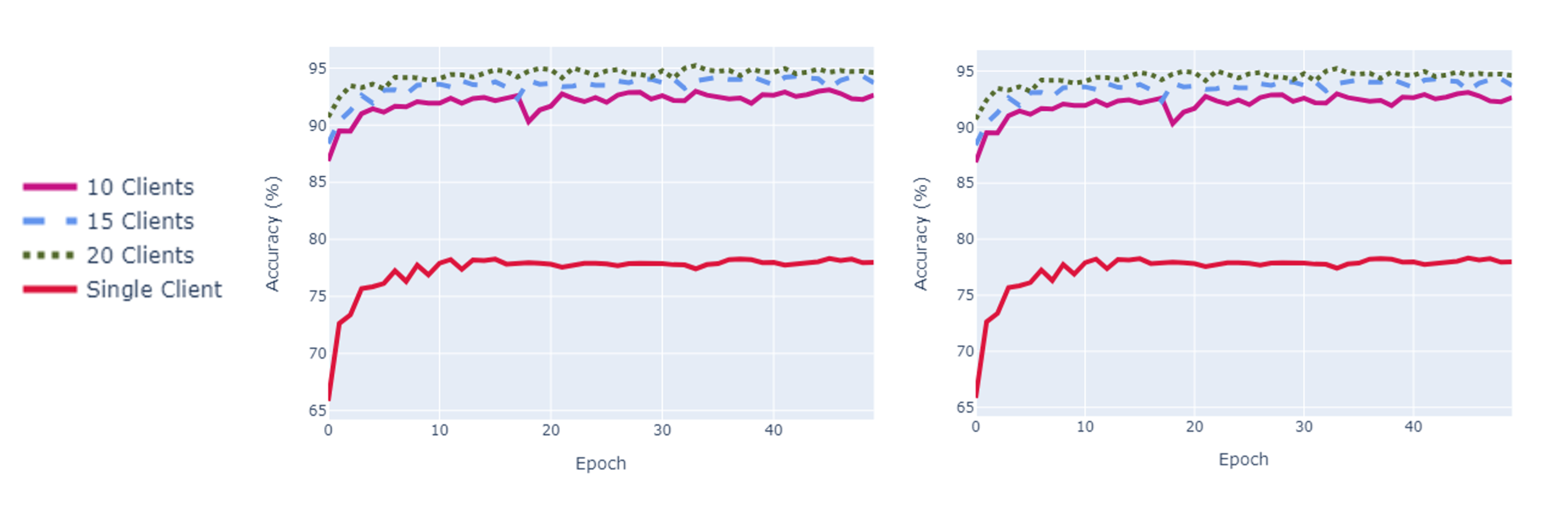}
  \caption{Centralised training with data from all the clients.} \label{fig:centraldata}
\end{figure*}

Plotly \cite{plotly}, a tool for data analysis and visualisation, was used for the creation of all the graphs. The results obtained for the tests with centralised data can be seen in \autoref{fig:centraldata}. The maximum efficiency achieved by a single client is \textbf{78.23\%}, which is not a high number, but it is reasonable since it does not have enough samples to achieve a higher effectiveness. Furthermore, from epoch 15 on-wards the loss starts to increase, which may be due to overfitting as the model has few samples. 

\begin{figure*}[h]
 \includegraphics[width=\textwidth]{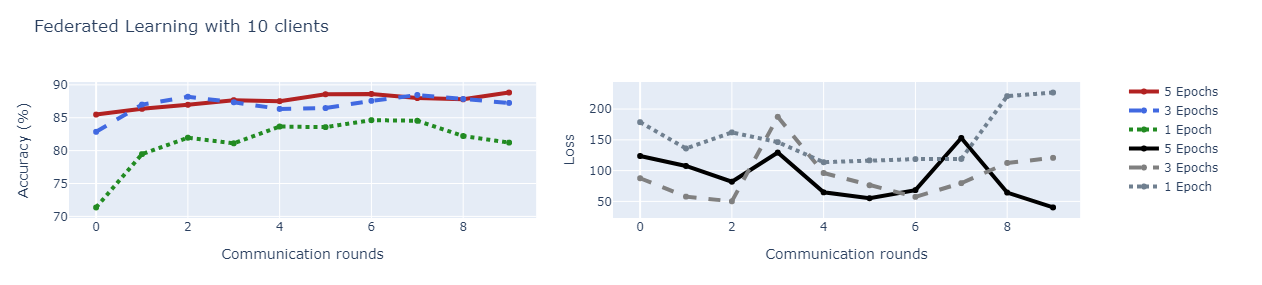}
  \includegraphics[width=\textwidth]{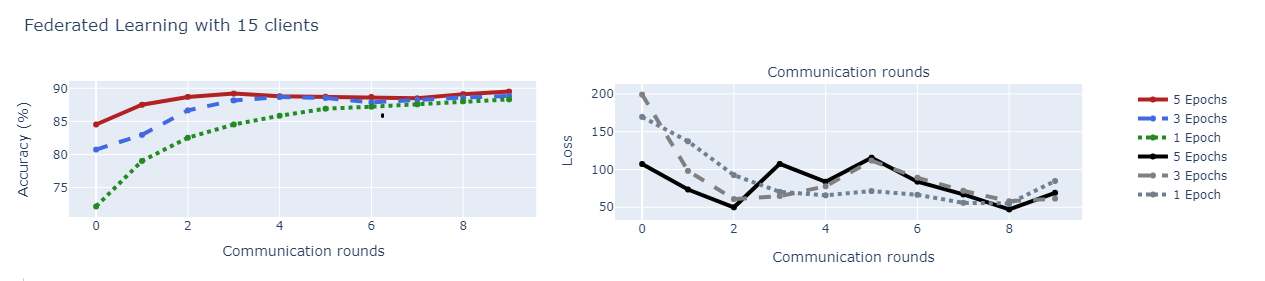}
   \includegraphics[width=\textwidth]{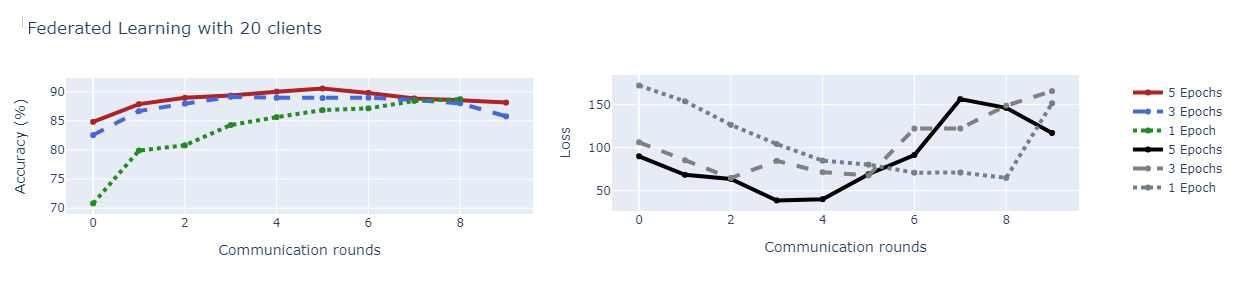}
 \caption{Federated Learning results.} \label{fig:FLdata}
\end{figure*}

In contrast, the results with data from 20, 10 and 15 clients are much better. These have respectively 3000, 4500 and 6000 samples for training, and achieve efficiencies of \textbf{92.87\%}, \textbf{94.26\%} and \textbf{95.23\%}. In these cases, the loss starts to increase around epoch 20-25, training from then on does not improve the model.

In the case of the Federated Learning tests, 10, 15 and 20 clients were also used. In turn, for each of these cases, 1,3,5 epochs were used, i.e. the number of times the model will be trained on the entire dataset. This is to test how it would influence the fact that more training would be done on all models in each round of communication. The results can be found in \autoref{fig:FLdata}.

It can be noticed how effectiveness improves as the number of clients and the amount of training in each round of communication increases. Since effectiveness improves with the number of epochs, a compromise should be found with the effort that each client makes to train his model. A full breakdown of the results can be found in \autoref{tab:comparationgraph}

\begin{table}[h] \centering
\begin{tabular}{ |l|l|l|l| }
 \hline    & \hfil 10 Clients & \hfil 15 Clients & \hfil 20 Clients \\
 \hline  \hfil 1 Epoch F.L. & \hfil 84.63\% & \hfil 88.03\% & \hfil 88.42\% \\
 \hline  \hfil 3 Epoch F.L.  & \hfil 88.03\%  & \hfil 88.68\% & \hfil 89.13\% \\
 \hline   \hfil 5 Epoch F.L. & \hfil 88.73\% & \hfil 89.2\% & \hfil 90.55\% \\
 \hline   \hfil Centralized data  & \hfil 92.87\%  & \hfil 94.26\% & \hfil 95.23\% \\
 \hline
\end{tabular} \caption{Accuracy Comparation}\label{tab:comparationgraph}
\end{table}

 \begin{figure*}[h]  \centering
 \includegraphics[width=\textwidth]{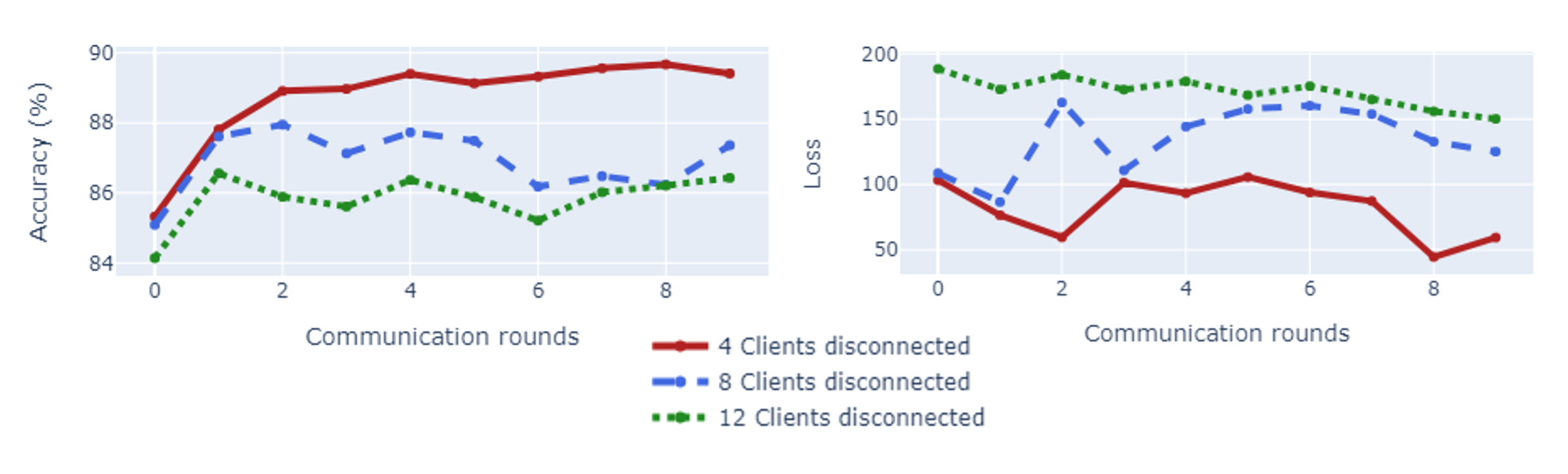}
  \caption{Performance in a hostile environment.} \label{fig:hostile}
\end{figure*}

The highest number achieved with Federated Learning is \textbf{90.55\%}, which is a respectable result, although it is a little far from the \textbf{95.23\%} that would be obtained with centralised data. However, although it sometimes happens, the objective of Federated Learning is not to improve the results of traditional centralised models, but to serve in cases where it is not possible to use them. Actually, even the worst result of the scenarios tested with Federated Learning, with 10 clients and 1 epoch, and \textbf{84.55\%} accuracy, is quite better than the \textbf{78.23\%} achieved by an single client.

\subsection{Hostile environment}
\label{subsec:hostile}
\medskip
Usual Federated Learning scenarios are unfriendly. As their use with mobile devices is common, many problems can arise, such as being disconnected due to lack of battery, or not having an internet connection.

The performance of this prototype was put to the test in this type of environment. The results were tested with the most effective set of the ones above, 20 clients and 5 epochs. A comparison was made if 4, 8 and 12 clients were disconnected in each round of communication. The results can be found in \autoref{fig:hostile}

When a client joins the communication, it waits until it receives the model from the server. This way, the clients pick up the most updated model when it joins. In addition, it gets the current round of communication. The fact that communication takes place with no need for the server to know which clients are interacting at any given moment makes it work well in this situation. 

In this case, the results are very influenced by which clients are disconnected in each round of communication. For example, it can be seen how for the intermediate scenario the loss fluctuates considerably. 

In terms of accuracy, values of \textbf{86.37\%}, \textbf{87.95\%} and \textbf{89.67\%}  were reached respectively. A comparison can be drawn between the case with hostile environment and 4 clients disconnecting in each round, and the friendly environment case where there were always 15 clients. The first case achieves a better result because in addition to 1 more client, after finishing the ten rounds of communication it has managed to take into account the data of all 20 clients, while the second case is limited to the data of the same 15 clients.

\section{Conclusions and future work}
\label{sec:conclusions}

This article explores the feasibility and effectiveness of using Federated Learning, concluding that the use of these techniques is highly beneficial. A prototype was implemented using an Amazon Web Services EC2 instance as the coordinator server and raspberry's boards as edge devices. In our experiments, Federated Learning has been proven to achieve better results than would be achieved by an individual client (edge device). However, it does not necessarily improve the performance of traditional techniques where all data is centralised.

 Moreover, its application is also useful in situations where the environment is not comfortable. One of its main advantages is that its implementation is still beneficial when communication is unstable or slow. These are the regular scenarios in the realisation of the IoT, especially where FL will be commonly deployed. Last but not least, a edge-based FL approach is also beneficial in terms of privacy and security as the data remains at the edge and only  model are exposure to inferential and poisoning model attacks among other.   In future lines of work, interesting paths remain to be explored. Some of the possibilities would be to test how performance would be improved by using more clients and larger datasets. This would allow to know how important it is to reach a certain number of customers. Another possibility is to use alternatives to FederatedAVG, such as CO-OP \cite{wang2017co}, which allows communication to be asynchronous. On the other hand, FL works only when data are local and the edge devices collaborate perfectly, thus being unsuitable to distributed training and inference. FL generally involves the exchange of large volumes of data, so a naive deployment over wireless networks is exceedingly costly, slow and vulnerable to outer attacks or to hidden collusion among the computing nodes. To overcome those fundamental bottlenecks, another line of future work focuses on distributed ML as the appropriate approach for addressing the performance issues and the privacy requirements that edge-intelligent services demand. 

\begin{acks}
This work was supported by the European Regional Development Fund (ERDF) and the Galician Regional Government, under the agreement for funding the Atlantic Research Center for Information and Communication Technologies (AtlanTTic). This work was also supported by the Spanish Government under research project “Enhancing Communication Protocols with Machine Learning while Protecting Sensitive Data (COMPROMISE) (PID2020-113795RB-C33/AEI/10.13039/501100011033).
\end{acks}

%\citestyle{acmauthoryear}
\bibliographystyle{ACM-Reference-Format}
\bibliography{myreferences}

\end{document}